\documentclass[10pt,twocolumn,letterpaper]{article}

\usepackage{cvpr}
\usepackage{times}
\usepackage{epsfig}
\usepackage{graphicx}
\usepackage{amsmath}
\usepackage{amssymb}
\usepackage{booktabs}
\usepackage{multirow}
\usepackage{subfigure}
\usepackage{lib}
\usepackage[font=small,labelfont=bf,skip=5pt]{caption}

\usepackage[pagebackref=true,breaklinks=true,letterpaper=true,colorlinks,bookmarks=false]{hyperref}

\cvprfinalcopy 

\def\cvprPaperID{722} 

\makeatletter
\def\blfootnote{\xdef\@thefnmark{}\@footnotetext}
\makeatother

\makeatletter
\renewcommand{\paragraph}{%
  \@startsection{paragraph}{4}%
  {\z@}{1.5ex \@plus 1ex \@minus .2ex}{-0.5em}%
  {\normalfont\normalsize\bfseries}%
}
\makeatother

\ifcvprfinal\fi
\begin{document}

\title{Finding Action Tubes}

\author{Georgia Gkioxari\\
UC Berkeley\\
{\tt\small gkioxari@eecs.berkeley.edu}
\and
Jitendra Malik\\
UC Berkeley\\
{\tt\small malik@eecs.berkeley.edu}
}

\maketitle

\begin{abstract}
 We address the problem of action detection in videos. Driven by the latest progress in object detection from 2D images, we build action models using rich feature hierarchies derived from shape and kinematic cues. We incorporate appearance and motion in two ways. First, starting from image region proposals we select those that are motion salient and thus are more likely to contain the action. This leads to a significant reduction in the number of regions being processed and allows for faster computations. Second, we extract spatio-temporal feature representations to build strong classifiers using Convolutional Neural Networks. We link our predictions to produce detections consistent in time, which we call \textit{action tubes}. We show that our approach outperforms other techniques in the task of action detection.
\end{abstract}
\section{Introduction}
\seclabel{intro}

In object recognition, there are two traditional problems: whole image classification, ``\textit{is there a chair in the image?}'', and object detection, ``\textit{is there a chair and where is it in the image?}''. The two problems have been quantified by the PASCAL Visual Object Challenge \cite{PASCAL-ijcv,PASCAL11} and more recently the ImageNet Challenge \cite{imagenet_cvpr09,ILSVRC12}. The focus has been on the object detection task due to its direct relationship to practical, real world applications. When we turn to the field of action recognition in videos, we find that most work is focused on video classification,``\textit{is there an action present in the video}'', with leading approaches \cite{wang2011,wang2013,simonyan2014} trying to classify the video as a whole. In this work,  we address the problem of action detection, ``\textit{is there an action and where is it in the video}''. 

Our goal is to build models which can localize and classify actions in video. Inspired by the recent advances in the field of object detection in images \cite{girshick2014rcnn}, we start by selecting candidate regions and use convolutional networks to classify them. Motion is a valuable cue for action recognition and we utilize it in two ways. We use motion saliency to eliminate regions that are not likely to contain the action. This leads to a big reduction in the number of regions being processed and subsequently in compute time. Additionally, we incorporate kinematic cues to build powerful models for action detection. \figref{overview} shows the design of our action models. Given a region, appearance and motion cues are used with the aid of convolutional neural networks (CNNs) to make a prediction. Our experiments indicate that appearance and motion are complementary sources of information and using both leads to significant improvement in performance (\secref{results}). Predictions from all the frames of the video are linked to produce consistent detections in time. We call the linked predictions in time \textit{action tubes}. \figref{approach} outlines our approach.

Our detection pipeline is inspired by the human vision system and, in particular, the two-streams hypothesis  \cite{GoodaleMilner92}. The ventral pathway (``\textit{what pathway}'') in the visual cortex responds to shape, color and texture while the dorsal pathway (``\textit{where pathway}'') responds to spatial transformations and movement. We use convolutional neural networks to computationally simulate the two pathways. The first network, \textit{spatial-CNN}, operates on static cues and captures the appearance of the actor and the environment. The second network, \textit{motion-CNN}, operates on motion cues and captures patterns of movement of the actor and the object (if any) involved in the action. Both networks are trained to discriminate between the actors and the background as well as between actors performing different actions. 

We show results on the task of action detection on two  publicly available datasets, that contain actions in real world scenarios, UCF Sports \cite{UCFsports} and J-HMDB \cite{J-HMDB}. These are the only datasets suitable for this task, unlike the task of action classification, where more datasets and of bigger size (up to 1M videos) exist.
Our approach outperforms all other approaches (\cite{Jain2014,WangQT14,SDPM,LanWM11}) on UCF sports, with the biggest gain observed for high overlap thresholds. In particular, for an overlap threshold of 0.6 our approach shows a relative improvement of 87.3\%, achieving mean AUC of 41.2\% compared to 22.0\% reported by \cite{WangQT14}. On the larger J-HMDB, we present an ablation study and show the effect of each component when considered separately. Unfortunately, no other approaches report numbers on this dataset. Additionally, we show that action tubes yield state-of-the-art results on action classification on J-HMDB. Using our action detections we are able to achieve an accuracy of 62.5\% on J-HMDB, compared to 56.6\% reported by \cite{wang2011}, the previous state-of-the-art approach on video classification.

The rest of the paper is organized as follows. In \secref{related} we mention related work on action classification and action detection in videos. In \secref{approach} we describe the details of our approach. In \secref{results} we report our results on the two datasets.

\begin{figure*}[t]
\begin{center}
  \includegraphics[width=0.8\linewidth]{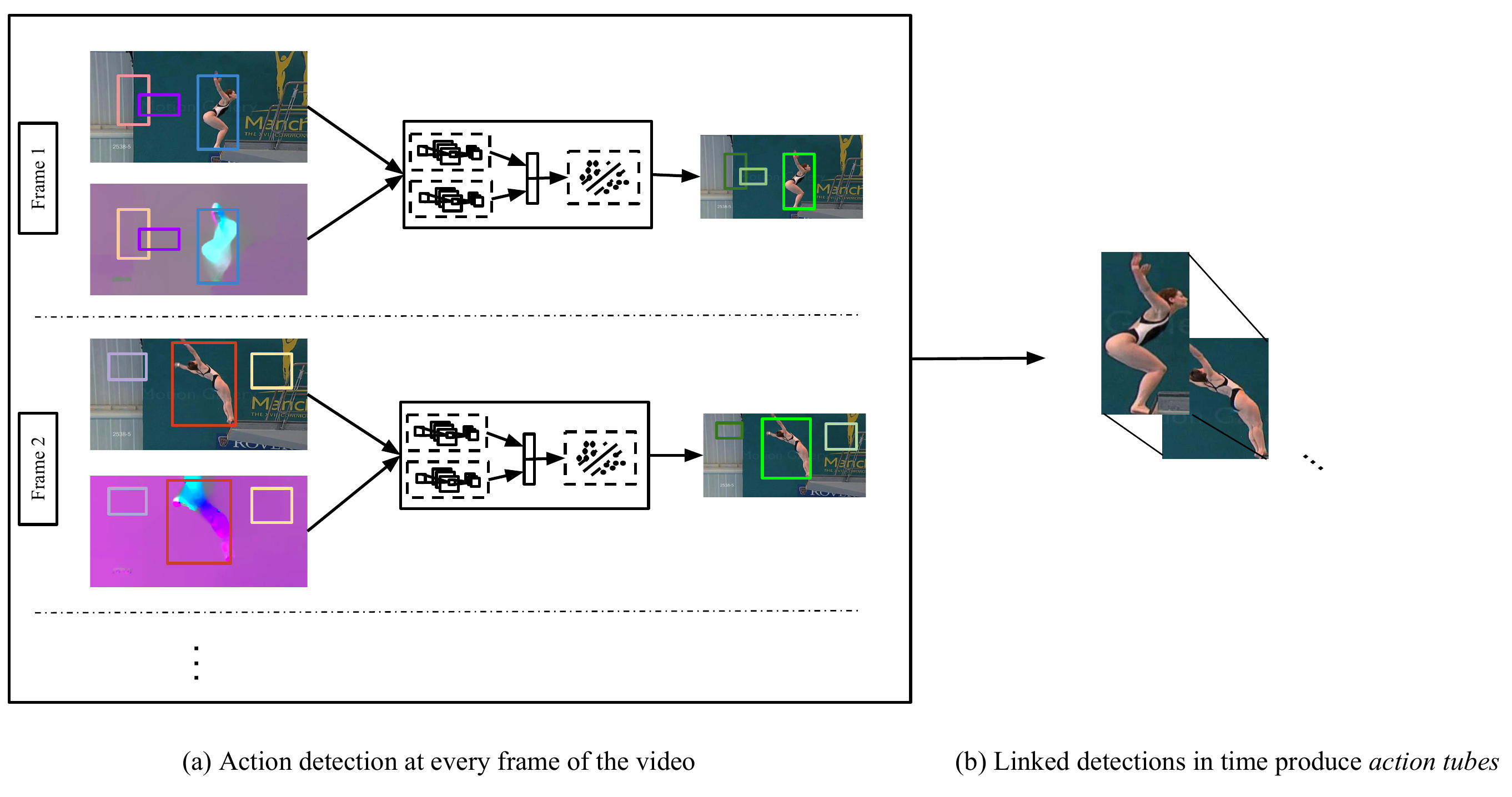}
\end{center}
   \caption{An outline of our approach. (a) Candidate regions are fed into action specific classifiers, which make predictions using static and motion cues. (b) The regions are linked across frames based on the action predictions and their spatial overlap. \textit{Action tubes} are produced for each action and each video.}
   \figlabel{approach}
\end{figure*}

\section{Related Work}
\seclabel{related}

There has been a fair amount of research on action recognition. We refer to \cite{AggarwalActivity,PoppeSurvey,WeinlandSurvey} for recent surveys in the field. For the task of action classification, recent approaches use features based on shape (e.g. HOG \cite{Dalal05}, SIFT \cite{SIFT}) and motion (e.g. optical flow, MBH \cite{Dalal2006}) with high order encodings (e.g. Bag of Words, Fischer vectors) and train classifiers (e.g. SVM, decision forests)  to make action predictions. More specifically, Laptev \etal \cite{LMSR08} extract local features at spatio-temporal interest points which they encode using Bag of Words and train SVM classifiers. Wang \etal \cite{wang2011} use dense point trajectories, where features are extracted from regions which are being tracked using optical flow across the frames, instead of fixed locations on a grid space. Recently, the authors improved their approach \cite{wang2013} using camera motion to correct the trajectories. They estimate the camera movement by matching points between frames using shape and motion cues after discarding those that belong to the humans in the frame. The big relative improvement of their approach shows that camera motion has a significant impact on the final predictions, especially when dealing with real world video data. Jain \etal \cite{Jain2013} make a similar observation.

Following the impressive results of deep architectures, such as CNNs, on the task of handwritten digit recognition \cite{lecun-89e} and more recently image classification \cite{krizhevsky2012imagenet} and object detection in images \cite{girshick2014rcnn}, attempts have been made to train deep networks for the task of action classification. Jhuang \etal \cite{Jhuang2007} build a feedforward network which consists of a hierarchy of spatio-temporal feature detectors of predesigned motion and shape filters, inspired by the dorsal stream of the visual cortex. Taylor \etal \cite{taylor2010} use convolutional gated RBMs to learn features for video data in an unsupervised manner and apply them for the task of action classification. More recently, Ji \etal \cite{3Dconv} build 3D CNNs, where convolutions are performed in 3D feature maps from both spatial and temporal dimensions. Karpathy \etal \cite{karpathy2014} explore a variety of network architectures to tackle the task of action classification on 1M videos. They show that operating on single frames performs equally well than when considering sequences of frames. Simonyan \& Zisserman \cite{simonyan2014} train two separate CNNs to explicitly capture spatial and temporal features. The spatial stream operates on the RGB image while the temporal stream on the optical flow signal. The two stream structure in our network for action detection is similar to their work, but the crucial difference is that their network is for full image classification while our system works on candidate regions and can thus localize the action. Also, the way we do temporal integration is quite different since our work tackles a different problem.

Approaches designed for the task of action classification use feature representations that discard any information regarding the location of the action. However, there are older approaches which are figure centric. Efros \etal \cite{30pixman} combine shape and motion features to build detectors suitable for action recognition at low resolution and predict the action using nearest neighbor techniques, but they assume that the actor has already been localized. Sch\"{u}ldt \etal \cite{Laptev2004} build local space-time features to recognize action patters using SVM classifiers. Blank \etal \cite{Irani2005} use spatio-temporal volume silhouettes to describe an action assuming in addition known background. More recently, per-frame human detectors have been used. Prest \etal \cite{prest:hal-00720847} propose to detect humans and objects and then model their interaction. Lan \etal \cite{LanWM11} learn spatio-temporal models for actions using figure-centric visual word representation, where the location of the subject is treated as a latent variable and is inferred jointly with the action label. Raptis \etal \cite{RaptisKS12} extract clusters of trajectories and group them to predict an action class using a graphical model. Tian \etal \cite{SDPM} extend the deformable parts model, introduced by \cite{lsvm-pami} for object detection in 2D images, to video using HOG3D feature descriptors \cite{klaser:inria-00514845}. Ma \etal extract segments of the human body and its parts based on color cues, which they prune using motion and shape cues. These parts serve as regions of interest from which features are extracted and subsequently are encoded using Bag of Words. Jain \etal \cite{Jain2014} produce space-time bounding boxes, starting from super-voxels, and use motion features with Bag of Words to classify the action within each candidate. Wang \etal \cite{WangQT14} propose a unified approach to discover effective action parts using dynamical poselets and model their relations.

\section{Building action detection models}
\seclabel{approach}

\figref{approach} outlines our approach. We classify region proposals using static and kinematic cues (stage a). The classifiers are comprised of two Convolutional Neural Networks (CNNs) which operate on the RGB and flow signal respectively. We make a prediction after using action specific SVM classifiers trained on the spatio-temporal representations produced by the two CNNs. We link the outputs of the classifiers across the frames of the videos (stage b) to produce \textit{action tubes}. 

\subsection{Regions of interest}

Given a frame, the number of possible regions that contain the action is enormous. However, the majority of these candidates are not descriptive and can be eliminated without loss in performance. There has been a lot of work on generating useful region proposals based on color, texture, edge cues (\cite{UijlingsIJCV2013, APBMM2014}). We use selective search \cite{UijlingsIJCV2013} on the RGB frames to generate approximately 2K regions per frame. Given that our task is to localize the actor, we discard the regions that are void of motion, using the optical flow signal. As a result, the final regions we consider are those that are salient in shape and motion.

Our motion saliency algorithm is extremely simple. We view the normalized magnitude of the optical flow signal $f_m$ as a heat map at the pixel level. If $R$ is a region, then $f_m(R)=\frac{1}{|R|}\sum_{i \in R} f_m(i)$ is a measure of how motion salient $R$ is. $R$ is discarded if $f_m(R)<\alpha$.

For $\alpha = 0.3$, approximately 85\% of boxes are discarded, with a loss of only 4\% in recall on J-HMDB, for an overlap threshold of 0.5. Despite the small loss in recall, this step is of great importance regarding the algorithm's time complexity. To be precise, it takes approximately 11s to process an image with 2K boxes, with the majority of the time being consumed in extracting features for the boxes (for more details see \cite{girshick2014rcnn}). This means that a video of 100 frames would require 18min to process! This is prohibitive, especially for a dataset of thousands of videos. Eliminating regions which are unlikely to contain the action reduces the compute time significantly.

\subsection{Action specific classifiers}

We use discriminative action classifiers on spatio-temporal features to make predictions for each region. The features are extracted from the final layer of the CNNs which are trained to discriminate among different actions as well as between actions and the background. We use a linear SVM with hard negative mining to train the final classifiers. \figref{overview} shows how spatial and motion cues are combined and fed into the SVM classifier.

\begin{figure}[t]
\begin{center}
  \includegraphics[width=0.8\linewidth]{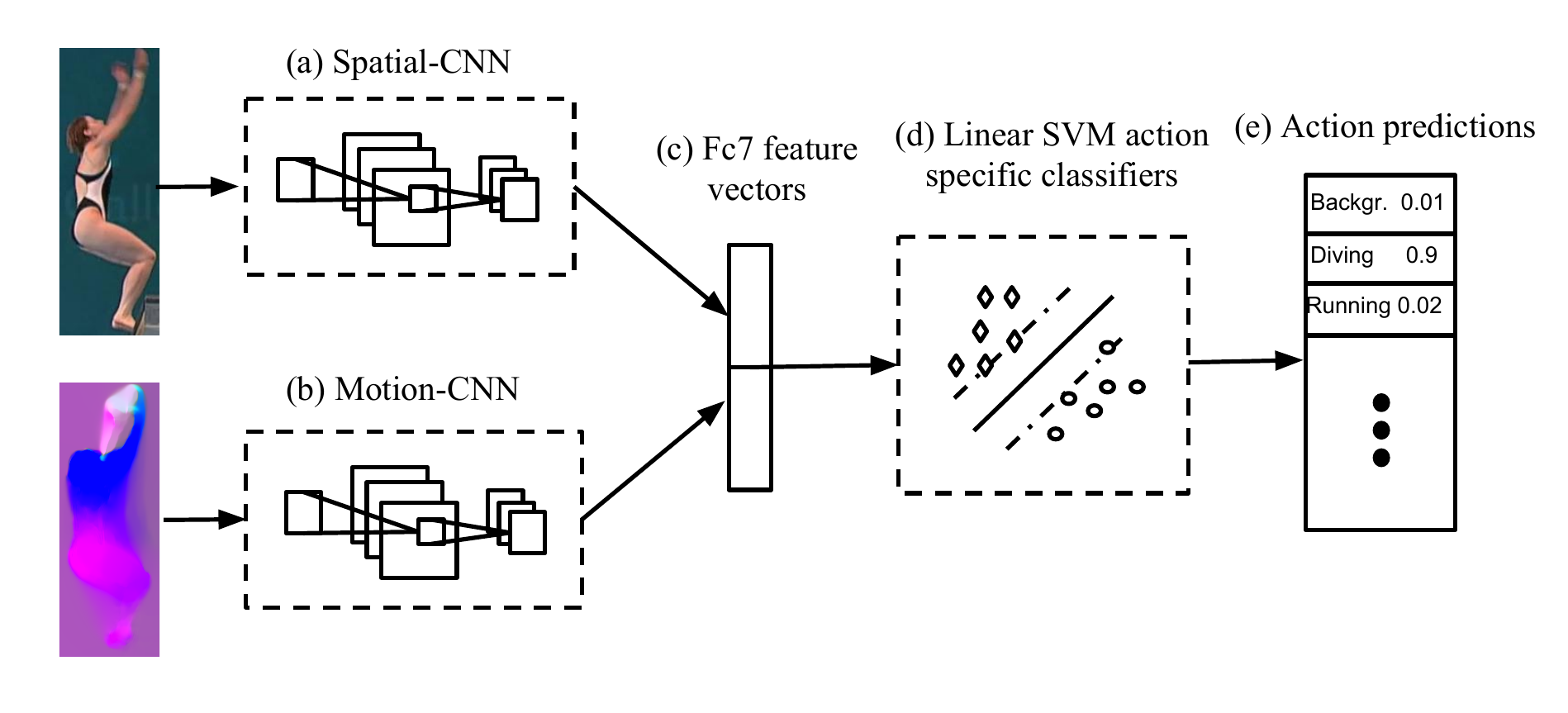}
\end{center}
   \caption{We use action specific SVM classifiers on spatio-temporal features. The features are extracted from the fc7 layer of two CNNs, \textit{spatial-CNN} and \textit{motion-CNN}, which were trained to detect actions using static and motion cues, respectively.}
   \figlabel{overview}
\end{figure}

\subsubsection{CNNs for action detection} 

We train two Convolutional Neural Networks for the task of action detection. The first network, \textit{spatial-CNN}, takes as input RGB frames and captures the appearance of the actor as well as cues from the scene. The second network, \textit{motion-CNN}, operates on the optical flow signal and captures the movement of the actor. Spatio-temporal features are extracted by combining the output from the intermediate layers of the two networks. Action specific SVM classifiers are trained on the spatio-temporal features and are used to make predictions at the frame level. \figref{overview} schematically outlines the procedure. Subsequently, we link the detections in time to produce temporarily consistent action predictions, which we call \textit{action tubes}.

We train  spatial-CNN and motion-CNN similar to R-CNN \cite{girshick2014rcnn}. Regions of interest are computed at every frame of the video, as described above. At train time, the regions which overlap more than 50\% with the ground truth are considered as positive examples, and the rest are negatives. The networks are carefully initialized to avoid overfitting. 

The architecture of spatial-CNN and motion-CNN is identical and follows \cite{krizhevsky2012imagenet} and \cite{zeiler2014}. Assume $C(k,n,s)$ is a convolutional layer with kernel size $k\times k$, $n$ filters and a stride of $s$, $P(k,s)$ a max pooling layer of kernel size $k\times k$ and stride $s$, $N$ a normalization layer, $RL$ a rectified linear unit, $FC(n)$ a fully connected layer with $n$ filters and $D(r)$ a dropout layer with dropout ratio $r$. The architecture of our networks follows: $C(7,96,2)-RL-P(3,2)-N-C(5,384,2)-RL-P(3,2)-N-C(3,512,1)-RL-C(3,512,1)-RL-C(3,384,1)-RL-P(3,2)-FC(4096)-D(0.5)-FC(4096)-D(0.5)-FC(|A|+1)$. The final fully connected layer has number of outputs as many as the action classes plus one for the background class. During training a softmax loss layer is added at the end of the network.

\paragraph{Network details} The architecture of our CNNs is inspired by two different network designs, \cite{krizhevsky2012imagenet} and \cite{zeiler2014}. Our network achieves 17\% top-5 error on the ILSVRC-2012 validation set for the task of classification.

\paragraph{Weight initialization}

Proper initialization is a key for training CNNs, especially in the absence of data. 

\noindent{\textbf{spatial-CNN}:} We want spatial-CNN to accurately localize people performing actions in 2D frames. We initialize spatial-CNN with a model that was trained on the PASCAL VOC 2012 detection task, similar to \cite{girshick2014rcnn}. This model has learned feature representations necessary for accurately detecting people under various appearance and occlusion patterns, as proven by the high person detection AP reported on the VOC2012 test set.

\noindent{\textbf{motion-CNN}:} We want motion-CNN to capture motion patterns. We train a network on single frame optical flow images for the task of action classification. We use the UCF101 dataset (split 1) \cite{UCF101}, which contains 13320 videos of 101 different actions. Our single frame optical flow model achieves an accuracy of 72.2\% on split 1, similar to 73.9\% reported by \cite{simonyan2014}. The 1.7\% difference can be attributed to the differences in the network's architectures. Indeed, the network used in \cite{simonyan2014} yields 13.5\% top-5 error on the ILSVRC-2012 validation set, compared to the 17\% top-5 error achieved by our network. This model is used to initialize motion-CNN when trained on smaller datasets, such as UCF Sports and J-HMDB.

\paragraph{Processing of input data} 

We preprocess the input for each of the networks as follows

\noindent{\textbf{spatial-CNN}:} The RGB frames are cropped to the bounds of the regions of interest, with a padding of 16 pixels, which is added in each dimension. The average RGB values are subtracted from the patches. During training, the patches are randomly cropped to $227\times227$ size, and are flipped horizontally with a probability of 0.5. 

\noindent{\textbf{motion-CNN}:} We compute the optical flow signal for each frame, according to \cite{Brox2004}. We stack the flow in the x-, y-direction and the magnitude to form a 3-dimensional image, and scale it by a constant ($s=16$). During training, the patches are randomly cropped and flipped.

\paragraph{Parameters}

We train spatial-CNN and motion-CNN with backpropagation, using Caffe \cite{jia2014caffe}. We use a learning rate of 0.001, a momentum of 0.9 and a weight decay of 0.0005. We train the networks for 2K iterations. We observed more iterations were unnecessary, due to the good initialization of the networks. 

\subsubsection{Training action specific SVM classifiers}

We train action specific SVM classifiers on spatio-temporal features, which are extracted from an intermediate layer of the two networks. More precisely, given a region $R$, let $\phi_s(R)$ and $\phi_m(R)$ be the feature vectors computed after the 7th fully connected layer in spatial-CNN and motion-CNN respectively. We combine the two feature vectors $\phi(R) = [\phi_s(R)^T \hspace{1mm} \phi_m(R)^T]^T$ to obtain a spatio-temporal feature representation for $R$. We train SVM classifiers $\mathbf{w}_\alpha$ for each action $\alpha \in A$, where ground truth regions for $\alpha$ are considered as positive examples and regions that overlap less than 0.3 with the ground truth as negative. During training, we use hard negative mining.

At test time, each region $R$ is a associated with a score vector $score(R) = \{ \mathbf{w}_\alpha^T \phi(R): \alpha \in A\}$, where each entry is a measure of confidence that action $\alpha$ is performed within the region.

\subsection{Linking action detections}

Actions in videos are being performed over a period of time. Our approach makes decisions on a single frame level. In order to create temporally coherent detections, we link the results from our single frame approach into unified detections along time.

Assume two consecutive frames at times $t$ and $t+1$, respectively, and assume $R_t$ is a region at $t$ and $R_{t+1}$ at $t+1$. For an action $\alpha$, we define the linking score between those regions to be

\begin{equation}
s_\alpha (R_t,R_{t+1}) = \mathbf{w}_\alpha^T  \phi(R_t) + \mathbf{w}_\alpha^T  \phi(R_{t+1}) + \lambda \cdot ov(R_t,R_{t+1})
\eqlabel{eq1}
\end{equation}
where $ov(R,\hat{R})$ is the intersection-over-union of two regions $R$ and $\hat{R}$ and $\lambda$ is a scalar. In other words, two regions are strongly linked if their spatial extent significantly overlaps and if they score high under the action model.

For each action in the video, we seek the optimal path 

\begin{equation}
 \bar{R}^*_\alpha= \argmax_{\bar{R}}  \frac{1}{T} \sum_{t=1}^{T-1} s_\alpha (R_t,R_{t+1})
 \eqlabel{eq2}
\end{equation}
where $\bar{R}_\alpha=[R_1, R_2, ..., R_T]$ is the sequence of linked regions for action $\alpha$. We solve the above optimization problem using the Viterbi algorithm. After the optimal path is found, the regions in $\bar{R}^*_\alpha$ are removed from the set of regions and  \eqref{eq2} is solved again. This is repeated until the set of regions is empty. Each path from \eqref{eq2} is called an \textit{action tube}. The score of an action tube $\bar{R}_\alpha$ is defined as $S_\alpha(\bar{R}_\alpha) =  \frac{1}{T} \sum_{t=1}^{T-1} s_\alpha (R_t,R_{t+1})$.

\section{Results}
\seclabel{results}

We evaluate our approach on two widely used datasets, namely UCF Sports \cite{UCFsports} and J-HMDB \cite{J-HMDB}. On UCF sports we compare against other techniques and show substantial improvement from state-of-the-art approaches. We present an ablation study of our CNN-based approach and show results on action classification using our action tubes on J-HMDB, which is a substantially larger dataset than UCF Sports.

\paragraph{Datasets}  UCF Sports consists of 150 videos with 10 different actions. There are on average 10.3 videos per action for training, and 4.7 for testing \footnote{The split was proposed by \cite{LanWM11}}. J-HMDB contains about 900 videos of 21 different actions. The videos are extracted from the larger HMDB dataset \cite{HMDB51}, consisting of 51 actions. Contrary to J-HMDB, UCF Sports has been widely used by scientists for evaluation purposes. J-HMDB is more interesting and should receive much more attention than it has in the past.

\paragraph{Metrics.} To quantify our results, we report Average-Precision at a frame level, \textit{frame-AP}, and at the video level, \textit{video-AP}. We also plot ROC curves and measure AUC, a metric commonly used by other approaches. None of the AP metrics have been used by other methods on this task. However, we feel they are informative and provide a direct link between the tasks of action detection and object detection in images.

\begin{itemize}
\item{\bf{frame-AP}} measures the area under the precision-recall curve of the detections for each frame (similar to the PASCAL VOC detection challenge \cite{PASCAL-ijcv}). A detection is correct if the intersection-over-union with the ground truth at that frame is greater than $\sigma$ and the action label is correctly predicted.
\item{\bf{video-AP}} measures the area under the precision-recall curve of the action tubes predictions. A tube is correct if the mean per frame intersection-over-union with the ground truth across the frames of the video is greater than $\sigma$ and the action label is correctly predicted. 
\item{\bf{AUC}} measures the area under the ROC curve, a metric previously used on this task. An action tube is correct under the same conditions as in \textit{video-AP}. Following \cite{SDPM}, the ROC curve is plotted until a false positive rate of 0.6, while keeping the top-3 detections per class and per video. Consequently, the best possible AUC score is 60\%. 
\end{itemize}

\subsection{Results on UCF sports}

In \figref{ROC}, we plot the ROC curve for $\sigma = 0.2$ (red). In \figref{AUC} we plot the average AUC for different values of $\sigma$. We plot the curves as produced by the recent state-of-the-art approaches, Jain \etal \cite{Jain2014}, Wang \etal \cite{WangQT14}, Tian \etal \cite{SDPM} and Lan \etal \cite{LanWM11}. Our approach outperforms all other techniques by a significant margin for all values of $\sigma$, showing the most improvement for high values of overlap, where other approaches tend to perform poorly. In particular, for $\sigma = 0.6$, our approach achieves an average AUC of 41.2\% compared to 22.0\% by \cite{WangQT14}. 

\begin{figure}[t]
\begin{center}
  \includegraphics[width=0.8\linewidth]{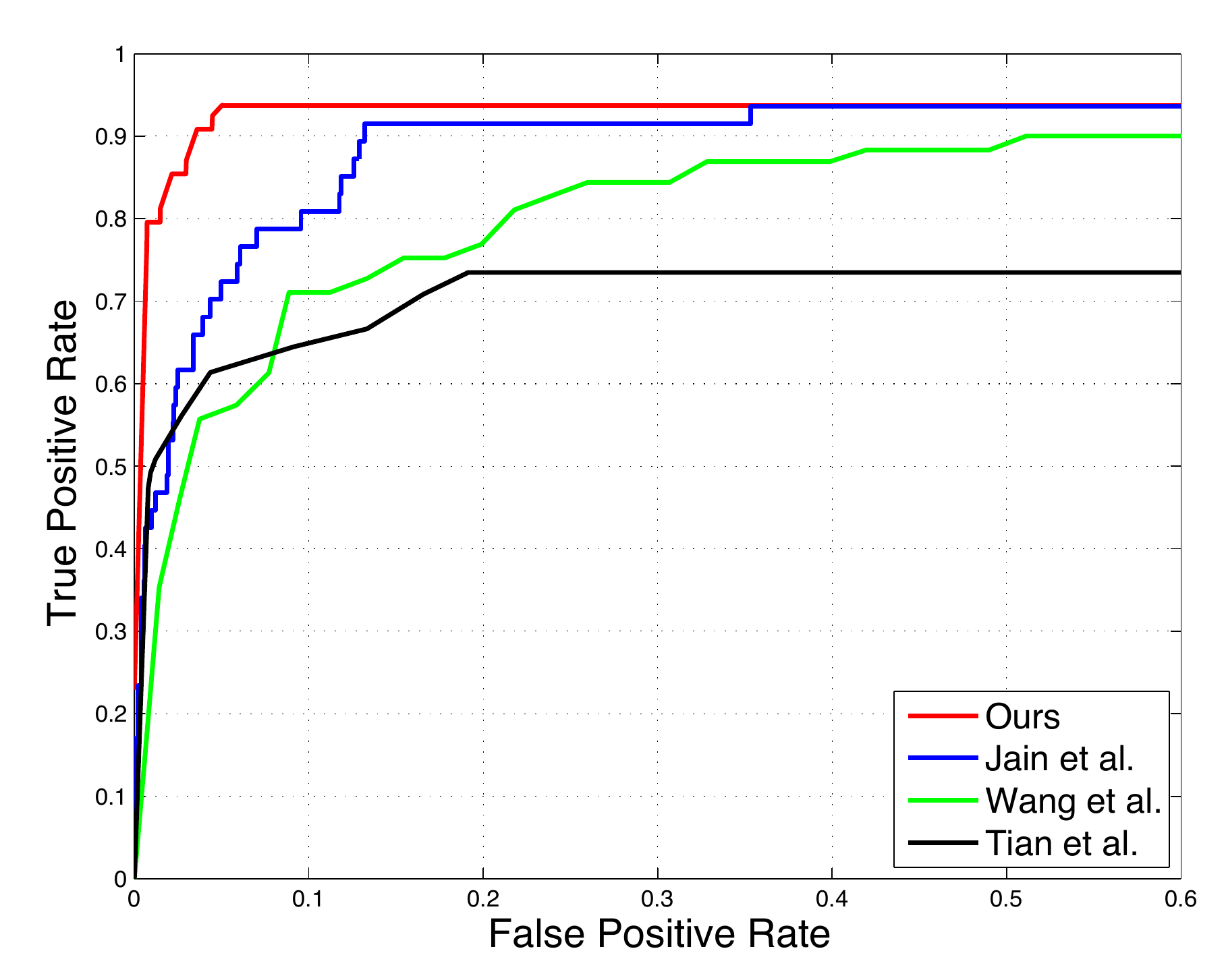}
\end{center}
   \caption{ROC curves on UCF Sports for an intersection-over-union threshold of $\sigma=0.2$. Red shows our approach. We manage to reach a high true positive rate at a much smaller false positive rate, compared to the other approaches shown on the plot.}
   \figlabel{ROC}
\end{figure}

\begin{figure}[t]
\begin{center}
  \includegraphics[width=0.8\linewidth]{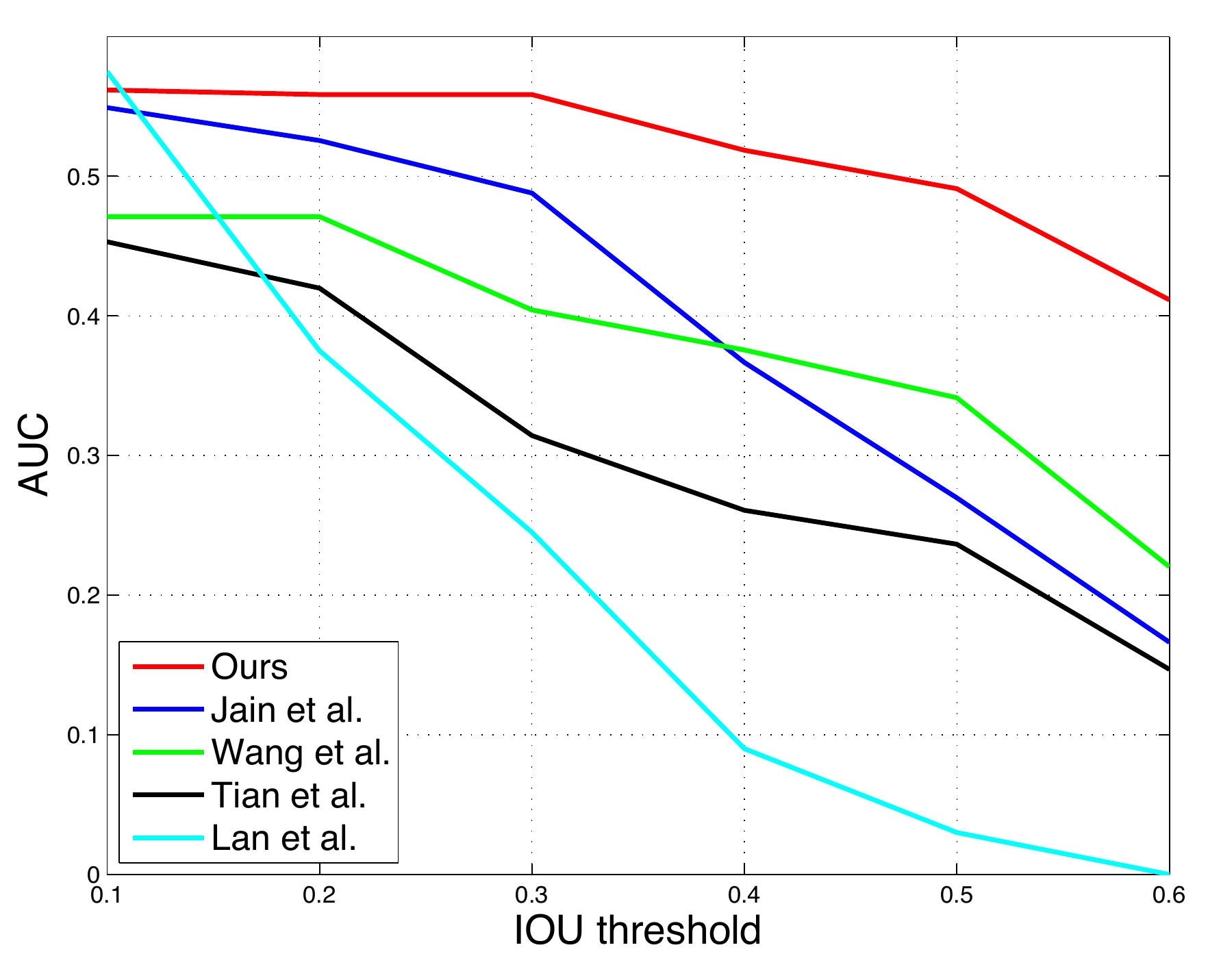}
\end{center}
   \caption{AUC on UCF Sports for various values of intersection-over-union threshold of $\sigma$ ($x$-axis). Red shows our approach. We consistently outperform other approaches, with the biggest improvement being achieved at high values of overlap ($\sigma \ge0.4$).}
   \figlabel{AUC}
\end{figure}

\tableref{UCF-AP} shows frame-AP (second row) and video-AP (third row) for an interestion-over-union threshold of $\sigma=0.5$. Our approach achieves a mean AP of 68.1\% at the frame level and 75.8\% at the video level, with excellent performance for most categories. \textit{Running} is the only action for which the action tubes fail to detect the actors (11.7 \% video-AP) , even though our approach is able to localize them at the frame level (54.9\% frame-AP).  This is due to the fact that the test videos for \textit{Running} contain multiple actors next to each other and our simple linking algorithm fails to consistently associate the detections with the correct actors, because of the proximity of the subjects and the presence of camera motion. In other words, the action tubes for \textit{Running} contain the action but the detections do not always correspond to the same person. Indeed, if we make our evaluation agnostic to the instance, video-AP for \textit{Running} is 83.8\%. Tracking objects in a video is a very interesting but rather orthogonal problem to action localization and is beyond the scope of this work.

\begin{table}[t!]
\centering
\renewcommand{\arraystretch}{1.2}
\renewcommand{\tabcolsep}{1.2mm}
\resizebox{\linewidth}{!}{
\begin{tabular}{@{}l|r*{9}{c}|cc@{}}
\textbf{AP (\%)}   & Diving  & Golf & Kicking & Lifting & Riding & Running  & Skateboarding & Swing1 & Swing2 & Walking & \emph{mAP} 
\\
\hline
frame-AP & 75.8 & 69.3 & 54.6 & 99.1 & 89.6 & 54.9 & 29.8 & 88.7 & 74.5 & 44.7 & 68.1\\
\hline \hline 
video-AP   & 100 & 91.7 & 66.7 & 100 & 100 & 11.7 & 41.7 & 100 & 100 & 45.8 & 75.8\\
\end{tabular}
}
\vspace{0.1em}
\caption{AP on the UCF Sports dataset for an intersection-over-union threshold of $\sigma=0.5$. \textit{frame-AP} measures AP of the action detections at the frame level, while \textit{video-AP} measures AP of the predicted action tubes.}
\tablelabel{UCF-AP}
\vspace{-0.5em}
\end{table}

\figref{UCFexamples} shows examples of detected action tubes on UCF sports. Each block corresponds to a different video. The videos were selected from the test set. We show the highest scoring action tube for each video. Red boxes indicate the detections in the corresponding frames. The predicted label is overlaid.

\subsection{Results on J-HMDB}

We report frame-AP and video-AP for the 21 actions of J-HMDB. We present an ablation study of our approach by evaluating the performance of the two networks,  \textit{spatial-CNN} and \textit{motion-CNN}. \tableref{JHMDB-AP} shows the results for each method and for each action category. 

As shown in the ablation study, it is apparent that the combination of spatial and motion-CNN performs significantly better for almost all actions. In addition, we can make some very useful observations. There are specific categories for which one signal matters more than the other. In particular, motion seems to be the most important for actions such as \textit{Clap}, \textit{Climb Stairs}, \textit{Sit}, \textit{Stand} and \textit{Swing Baseball}, while appearance contributes more for actions such as \textit{Catch}, \textit{Shoot Gun} and \textit{Throw}. Also, we notice that even though motion-CNN performs on average a bit worse than spatial-CNN at the frame level (24.3\% vs. 27.0\% respectively), it performs significantly better at the video level (45.7\% vs. 37.9\% respectively). This is due to the fact that the flow frames are not very informative when considered separately, however they produce a stronger overall prediction after the temporal smoothing provided by our linking algorithm.

\begin{table*}[t!]
\centering
\renewcommand{\arraystretch}{1.2}
\renewcommand{\tabcolsep}{1.2mm}
\resizebox{\linewidth}{!}{
\begin{tabular}{@{}l|r*{20}{c}|cc@{}}
\textbf{frame-AP (\%)}  &  brush\_hair & catch & clap & climb\_stairs & golf & jump & kick\_ball & pick & pour & pullup & push & run & shoot\_ball & shoot\_bow & shoot\_gun & sit & stand & swing\_baseball & throw & walk & wave & \emph{mAP} \\
\hline
spatial-CNN & 55.8 & \bf{25.5} & 25.1 & 24.0 & 77.5 & \phz1.9 & \phz5.3 & 21.4 & 68.6 & 71.0 & 15.4 & \phz6.3 & \phz4.6 & 41.1 & \bf{28.0} & \phz9.4 & \phz8.2 & 19.9 & \bf{17.8} & 29.2 & 11.5 & 27.0 \\
\hline
motion-CNN & 32.3 & \phz5.0 & 35.6 & 30.1 & 58.0 & \bf{\phz7.8} & \phz2.6 & 16.4 & 55.0 & 72.3 & \phz8.5 & \phz6.1 & \phz3.9 & 47.8 & \phz7.3 & 24.9 & 26.3 & \bf{36.3} & \phz4.5 & 22.1 & 7.6 & 24.3 \\
\hline
full & \bf{65.2} & 18.3 & \bf{38.1} & \bf{39.0} & \bf{79.4} & \phz7.3 & \bf{\phz9.4} & \bf{25.2} & \bf{80.2} & \bf{82.8} & \bf{33.6} & \bf{11.6} & \bf{\phz5.6} & \bf{66.8} & 27.0 & \bf{32.1} & \bf{34.2} & 33.6 & 15.5 & \bf{34.0} & \bf{21.9} & \bf{36.2} \\
\hline \hline 
\textbf{video-AP (\%)} &  \\
\hline
spatial-CNN  & 67.1 & \bf{34.4} & 37.2 & 36.3 & 93.8 & \phz7.3 & 14.4 & 29.6 & 80.2 & 93.9 & 17.4 & 10.0 & \phz8.8 & 71.2 & \bf{45.8} & 17.7 & 11.6 & 38.5 & 20.4 & 40.5 & 19.4 & 37.9\\
\hline
motion-CNN  & 66.3 & 16.0 & \bf{60.0} & 51.6 & 88.6 & \bf{18.9} & 10.8 & 23.9 & 83.4 & 96.7 & 18.2 & 17.2 & \bf{14.0} & 84.4 & 19.3 & \bf{72.6} & \bf{61.8} & \bf{76.8} & 17.3 & 46.7 & 14.3 & 45.7\\
\hline
full  & \bf{79.1} & 33.4 & 53.9 & \bf{60.3} & \bf{99.3} & 18.4 & \bf{26.2} & \bf{42.0} & \bf{92.8} & \bf{98.1} & \bf{29.6} & \bf{24.6} & 13.7 & \bf{92.9} & 42.3 & 67.2 & 57.6 & 66.5 & \bf{27.9} & \bf{58.9} & \bf{35.8} & \bf{53.3}\\
\end{tabular}
}
\vspace{0.1em}
\caption{Results and ablation study on J-HMDB (averaged over the three splits). We report \textit{frame-AP} (top) and \textit{video-AP} (bottom) for the spatial and motion component and their combination (full). The combination of the spatial- and motion-CNN performs significantly better under both metrics, showing the significance of static and motion cues for the task of action recognition.}
\tablelabel{JHMDB-AP}
\vspace{-0.5em}
\end{table*}

\figref{auc_jhmdb} shows the AUC for different values of the intersection-over-union threshold, averaged over the three splits on J-HMDB. Unfortunately, comparison with other approaches is not possible on this dataset, since no other approaches report numbers or have source code available.

\begin{figure}[t]
\begin{center}
  \includegraphics[width=0.8\linewidth]{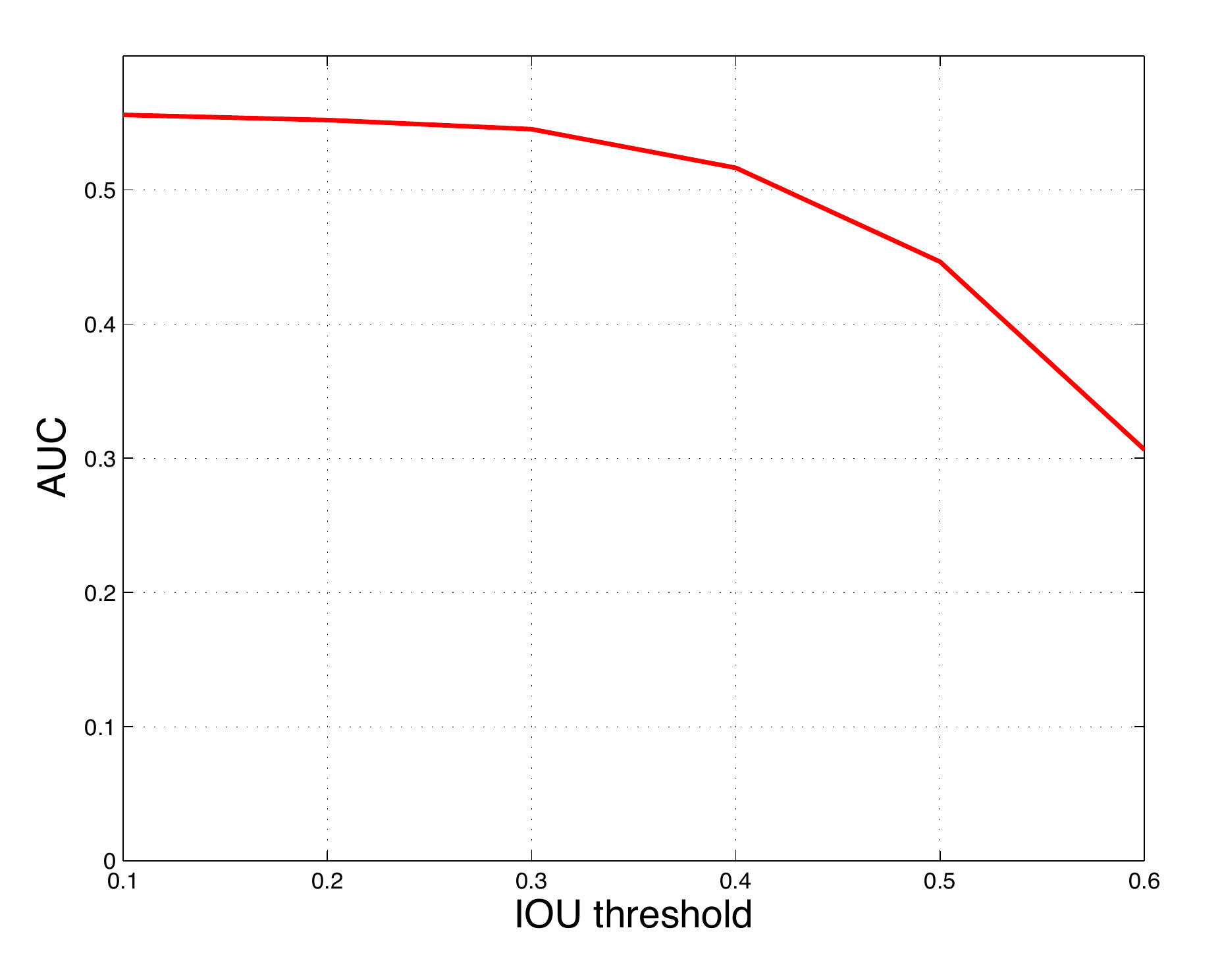}
\end{center}
   \caption{AUC on J-HMDB for different values of intersection-over-union threshold (averaged over the three splits).}
   \figlabel{auc_jhmdb}
\end{figure}

\figref{JHMDBexamples} shows examples of action tubes on J-HMDB. Each block corresponds to a different video. The videos are selected from the split 1 test set. We show the highest scoring action tube for each video. Red boxes indicate the detections in the corresponding frames. The predicted label is overlaid.

\paragraph{Action Classification}
Our approach is not limited to action detection. We can use the action tubes to predict an action label for the whole video. In particular, we can predict the label $l$ for a video by picking the action with the maximum action tube score
\begin{equation}
l = \argmax_{\alpha \in A} \max_{\bar{R} \in \{\bar{R}_\alpha\}} S_\alpha(\bar{R})
\eqlabel{videolabel}
\end{equation}
where $S_\alpha(\bar{R})$ is the score of the action tube $\bar{R}$ as defined by \eqref{eq2}.

If we use \eqref{videolabel} as the prediction, our approach yields an accuracy of 62.5\%, averaged over the three splits of J-HMDB. \figref{ConfMatrix} shows the confusion matrix.

\begin{figure}[t]
\begin{center}
  \includegraphics[width=1\linewidth]{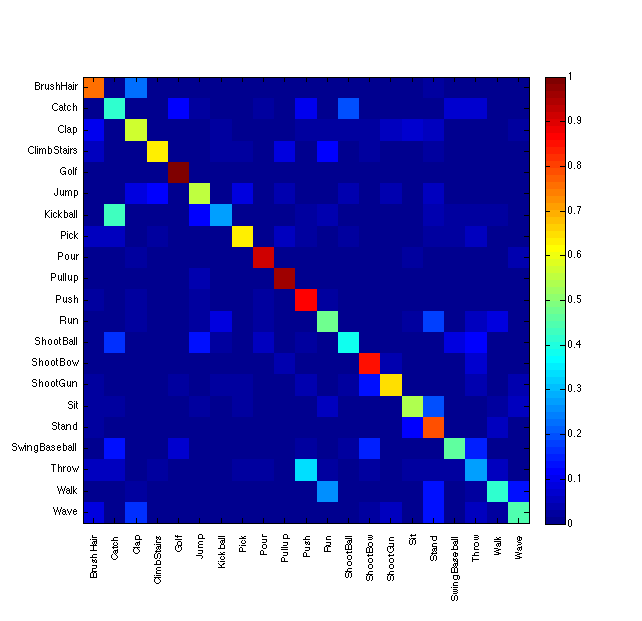}
\end{center}
   \caption{The confusion matrix on J-HMDB for the classification task, when using action tubes to predict a label for each video.}
   \figlabel{ConfMatrix}
\end{figure}

In order to show the impact of the action tubes in the above result, we adapt our approach for the task of action classification. We use spatial and motion-CNNs in a classification setting, where full frames are used as input instead of regions. The weights of the CNNs are initialized from networks trained on UCF 101 (split1) for the task of action classification. We average the class probabilities as produced by the softmax layers of the CNNs, instead of training SVM classifiers (We observed major overfitting problems when training SVM classifiers on top of the combined fc7 features). We average the outputs of spatial- and motion-CNNs, with weights 1/3 and 2/3 respectively, and pick the action label with the maximum score after averaging the frames of the videos. Note that our pipeline for classification is similar to \cite{simonyan2014}. This approach yields an accuracy of 56.5\% averaged over the three splits of J-HMDB. This compares to 56.6\% achieved by the  state-of-the-art approach \cite{wang2011}. \tableref{Classification} summarizes the results for action classification on J-HMDB. It is quite evident that focusing on the actor is beneficial for the task of video classification, while a lot of information is being lost when the whole scene is analyzed in an orderless fashion.

\begin{table}[t!]
\centering
\renewcommand{\arraystretch}{1.2}
\renewcommand{\tabcolsep}{1.2mm}
\resizebox{\linewidth}{!}{
\begin{tabular}{@{}l|r*{20}{c}|cc@{}}
\textbf{Accuracy (\%)}  &  Wang \etal \cite{wang2011} & CNN (1/3 spatial, 2/3 motion) & Action Tubes \\
\hline
J-HMDB & 56.6 & 56.5 & \bf{62.5} \\
\end{tabular}
}
\vspace{0.1em}
\caption{Classification accuracy on J-HMDB (averaged over the three splits). CNN (third column) shows the result of the weighted average of spatial and motion-CNN on the whole frames, while Action Tubes (fourth column) shows the result after using the scores of the predicted action tubes to make decisions for the video's label.}
\tablelabel{Classification}
\vspace{-0.5em}
\end{table}

\begin{figure*}[t]
\begin{center}
   \includegraphics[width=1\linewidth]{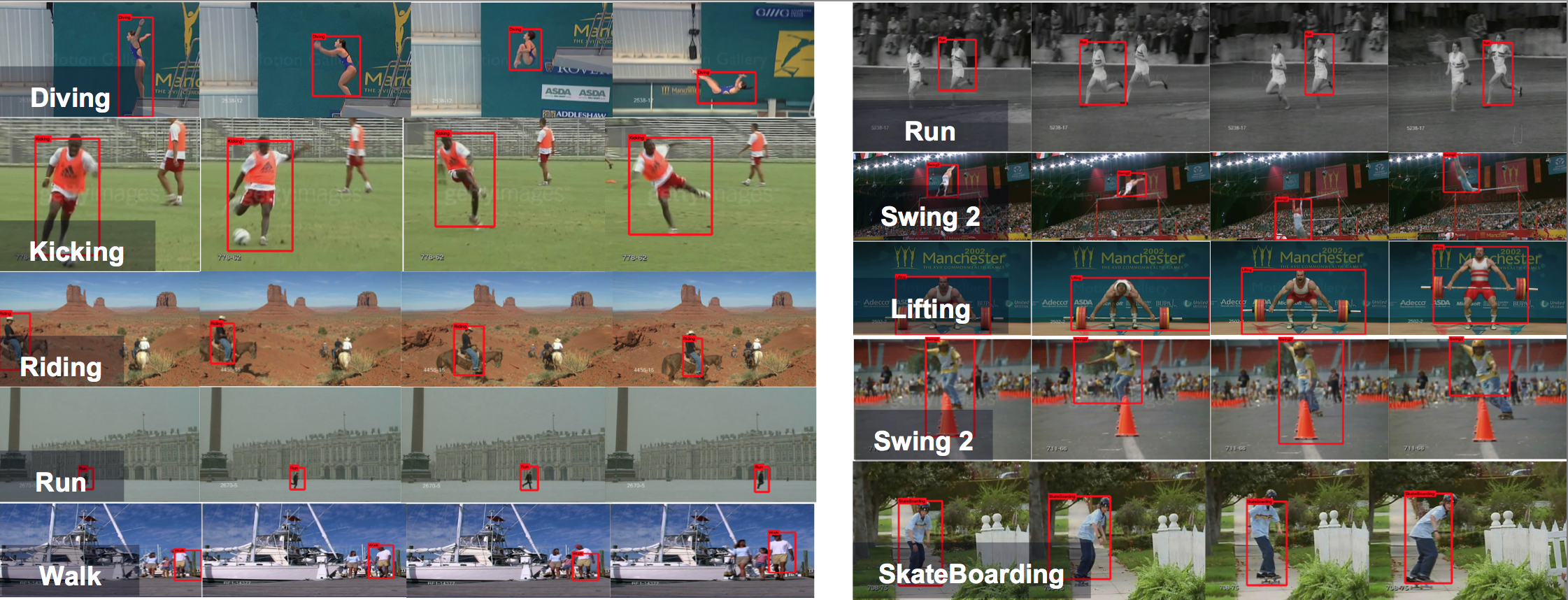}
\end{center}
   \caption{Examples from UCF Sports. Each block corresponds to a different video. We show the highest scoring action tube detected in the video. The red box indicates the region and the predicted label is overlaid. We show 4 frames from each video. The top example on the right shows the problem of tracking, while the 4th example on the right is a wrong prediction, with the true label being \textit{Skate Boarding}.} 
   \figlabel{UCFexamples}
\end{figure*}

\begin{figure*}[t]
\begin{center}
     \includegraphics[width=1\linewidth]{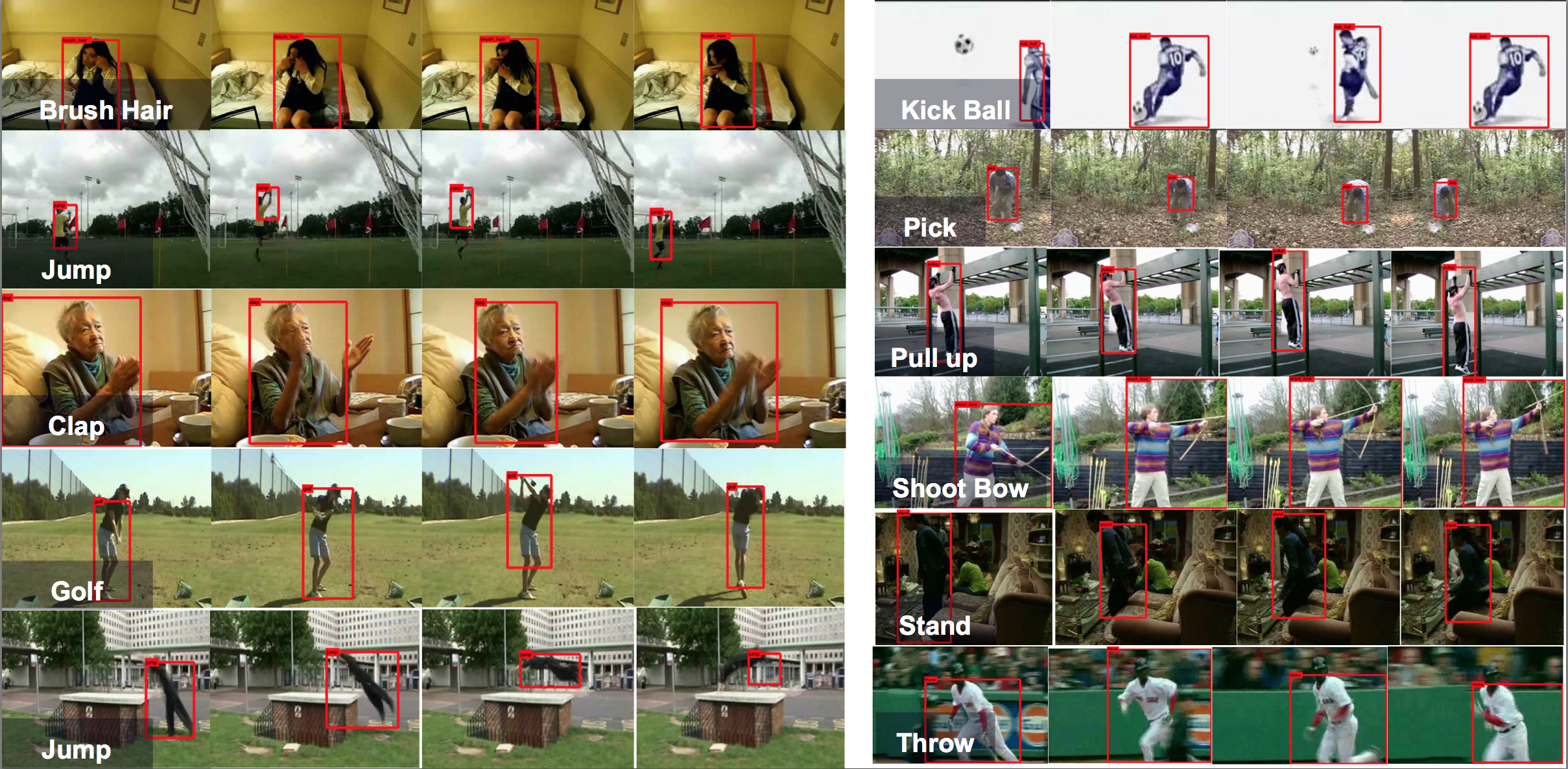}
\end{center}
   \caption{Examples from J-HMDB. Each block corresponds to a different video. We show the highest scoring action tube detected in the video. The red box indicates the region and the predicted label is overlaid. We show 4 frames from each video. The 2nd example on the left and the two bottom ones on the right are wrong predictions, with true labels being \textit{catch}, \textit{sit} and \textit{run} respectively.} 
   \figlabel{JHMDBexamples}
\end{figure*}

\section{Conclusions}

We propose an approach to action detection using convolutional neural networks on static and kinematic cues. We experimentally show that our action models perform state-of-the-art on the task of action localization. From our ablation study it is evident that appearance and motion cues are complementary and their combination is mandatory for accurate predictions across the board.

However, there are two problems closely related to action detection that we did not tackle. One is, as we mention in \secref{results}, the problem of tracking. For example, in a track field it is important to recognize that the athletes are running but also to be able to follow each one throughout the race. For this problem to be addressed, we need compelling datasets that contain videos of multiple actors, unlike the existing ones where the focus is on one or two actors. Second, camera motion is a factor which we did not examine, despite strong evidence that it has a significant impact on performance \cite{wang2013,Jain2013}. Efforts to eliminate the effect of camera movement, such as the one proposed by \cite{wang2013}, might further improve our results.


\clearpage
{\small
\bibliographystyle{ieee}
\bibliography{refs}
}

\end{document}